# A Very Compact Embedded CNN Processor Design Based on Logarithmic Computing

Tsung-Ying Lu, Hsu-Hsun Chin, Hsin-I Wu, and Ren-Song Tsay
Dept. of Computer Science
National Tsing Hua University, Taiwan
{iu0987810505, kim60138, hiwu.dery, rstsay}@gmail.com

## ABSTRACT

In this paper, we propose a very compact embedded CNN processor design based on a modified logarithmic computing method using very low bit-width representation. Our high-quality CNN processor can easily fit into edge devices. For Yolov2, our processing circuit takes only 0.15 mm2 using TSMC 40 nm cell library. The key idea is to constrain the activation and weight values of all layers uniformly to be within the range [-1, 1] and produce low bit-width logarithmic representation. With the uniform representations, we devise a unified, reusable CNN computing kernel and significantly reduce computing resources. The proposed approach has been extensively evaluated on many popular image classification CNN models (AlexNet, VGG16, and ResNet-18/34) and object detection models (Yolov2). The hardware-implemented results show that our design consumes only minimal computing and storage resources, yet attains very high accuracy. The design is thoroughly verified on FPGAs, and the SoC integration is underway with promising results. With extremely efficient resource and energy usage, our design is excellent for edge computing purposes.

## 1 Introduction

Deep convolutional neural networks (CNNs) have been successfully applied to many traditionally challenging analysis works, particularly visual imagery analysis [4][5]. However, CNN computations are known to be energy and storage space demanding. For instance, a sizeable single image using VGG16 [5] requires 15.5 billion MAC (multiply-accumulate) operations of 32-bit floating-point numbers and over 500 megabytes of memory space to store the *weight* parameters for convolutions. As a result, CNN remains to be a great challenge to deploy onto edge devices which have limited computing power and storage space. The edge devices generally refer to embedded devices, IoT (Internet-of-Things) devices, or low-end FPGA devices.

Although economic Binary-Weight-Networks and XNOR-Networks have been attempted [7], insufficient model accuracy curbs practical applications. Alternatively, low bit-width quantization approaches [3][6][9][10] are commonly adopted to reduce computation complexity by shrinking the length of the data representation. These approaches convert each CNN *weight* value into a low bit-width value. The reduced data bit-width generally cuts the storage space and computation complexity requirements. For instance, in theory, the 8-bit fixed-point AlexNet inferencing uses 75% fewer resources than the 32-bit fixed-point version, but with 2% model accuracy loss [8].

Then some others have proposed using logarithmic quantization methods for model accuracy improvement while having resource requirements similar to that of the fixed-point approach. The logarithmic quantization approaches quantize values onto the logarithmic domain. The logarithmic quantization requires fewer bits to represent a broader range of values, and hence many low bit-width logarithmic quantization approaches [11][17][25] claim to be able to produce better model accuracy than that of the fixed-point approaches. However, challenges emerge when applying the logarithmic quantization method on embedded systems or FPGAs

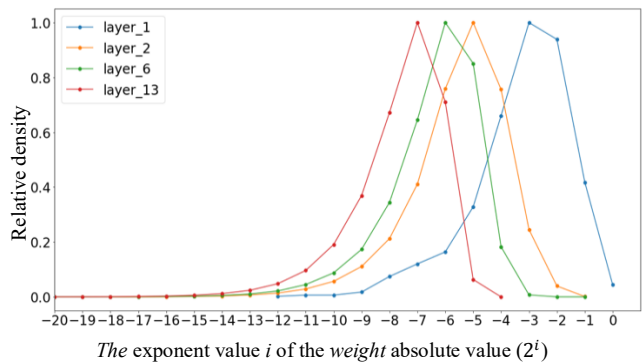

*The exponent value i of the weight absolute value ($2^i$)*

**Fig. 1**: The VGG16 example illustrates that each layer has different weight value range and wide bit-width is required to cover the whole range of weight values of all layers. The horizontal axis of the plot is the exponent value of the *weight* absolute value. The vertical axis is the relative density of the distributions. Each colored dashed-line pair covers the value range of each corresponding layer represented by the 4-bit logarithmic range. The grey area represents the 5-bit logarithmic range that covers 98% of all layer values.

due to a wide exponent value range, although the techniques apply conveniently on the CPUs or GPUs.

To illustrate the issue, we take a few layers from the VGG16 example shown in Fig. 1. Note that for convenience, all data are represented in absolute values since our focus is logarithmic quantization. If a 4-bit logarithmic representation is used, then the exponent value-ranges for layer-1 and layer-13 *weight* values would be [-13, -6] and [-8, -1] so that the value ranges may cover most data of the corresponding distributions, as indicated by the red and blue dashed-line pairs in Fig. 1. Note that one bit is excluded and used for the sign bit. Another solution is to use one more bit, i.e., 5 bits, to cover the grey area range [-15, 0], or 98% of all layer values. However, the approach consumes one more bit or equivalently more resource overheads to cover a wider value range and cause more computation and storage burden [2].

Motivated by the effectiveness of the batch normalization method [22][32], which regulates the input layer by adjusting and scaling the activations, we find that if the weight values of each layer can be normalized to the range of [-1, 1], then we can conveniently use the same low bit-width logarithmic representation without sacrificing accuracy. For instance, the 4-bit logarithmic representation is sufficient to cover the weight values of all layers after normalization.

This all-layer unified [-1, 1] normalization range approach is our key idea to produce a very compact CNN processor design. First, with the unified range, we can maintain low bit-width representation. Then since the computing structure of every layer is also unified, we can design a reusable processing kernel for all layers and dramatically save the silicon area. Additional silicon saving also comes from the fact that logarithmic computing requires only simple shift-and-add operations but no expensive multipliers.

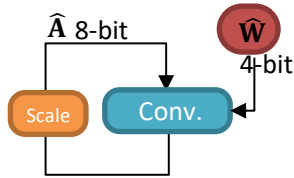

**Fig. 2**: The proposed unified CNN convolution layer architecture. **Scale** is a layer-specific scaling parameter to be executed after the convolution is complete. The *weight* and *activation* values are confined in [-1, 1] range. For actual implementation, the weights are in 4-bit logarithmic format and the activations are in 8-bit fixed-point format.

In practice, we observe that the input *activation* values are relatively uniform in distribution and are better represented in a fixed-point format but also normalized to [-1,1]. Then the only issue remains to streamline all layers is that the output data of each layer may not be in the range of [-1, 1]. To do this, we simply use a layer-specific scaling unit to normalize the output values back to be in the same range of [-1, 1]. In this way, the activation values of each layer will be consistently in the same value range.

With all layers having the same *weight* and *activation* value range, the convolution computation kernel implementation is very concise. Only a scaling unit after the convolution is needed to complete the unified architecture to achieve consistent and efficient computations with little overhead, as shown in Fig. 2 with no need for additional data-alignment units and extra *offset* parameters.

By confining the weigh and activation values to the range of [-1, 1], we may easily use 4-bit logarithmic represented weight values and 8-bit fixed-point activations to achieve almost the same model accuracy as that of the full-precision reference networks. Additionally, with a simple scaling unit, we can then reuse the same CNN computing kernel for all layers for minimum resource requirements. The actual implementations show surprisingly good quality results and prove that our design is very resource and energy-efficient and is excellent for edge computing purposes.

The organization of this paper is as follows. After reviewing the related work in Section 2, we describe our logarithmic computing approach in Section 3. Next, in Section 4, we present the experimental results. Finally, we conclude in Section 5.

## 2 Related Work

The work most related to our approach is the compression method. Different compression approaches have been proposed to reduce the resource requirement of CNNs for inferencing. In general, there are two categories of these approaches. The first is the network pruning approach [1][12][13][19][29], that prunes out ineffective *weight* computations. The second one is the network quantization approach, which includes floating-point [11], fixed-point [15][18][20] and logarithmic quantization [11][17][25] methods. Brief reviews of these approaches are presented below.

### 2.1 Network Pruning Approach

These methods tried to mitigate the over-parameterization [16] problem of CNNs by setting ineffective *weights* to zero values while attempting to retain model accuracy. For instance, LeCun et al. [19] computed the effect of each *weight* on the model accuracy based on the second-order approximation of the Taylor expansion for the network loss function. The low influence *weights* were removed, and the residual *weights* were retrained. This process was repeated until the desired *weight* reduction, and the acceptable model accuracy was reached. Practically, the second-order Taylor expansion requires a vast amount of computations, and hence, searching for low influence *weights* is very time-consuming.

Therefore, some developed heuristic approaches, which try to pick the redundant *weights* directly from the network and avoid elaborated search time. For instance, Han et al. [13] observed that *small weight* values have little effect on the layer's output. Hence they removed the *weight* values below an empirical threshold value and then retrained the residual *weights* to recover the model accuracy.

Although these network pruning approaches effectively reduce some *weights* to zero values, in reality, the zero values still require storage space. Therefore, the implementation overhead is inevitable.

Next, we discuss the quantization approach, which generally is more consistent in reducing model size.

### 2.2 Network Quantization Approach

Since CNNs may require hundreds of millions of *weight* parameters, which are usually in a 32-bit floating-point format, both the network storage and computing demand vast hardware resources. Therefore, some have proposed quantization approaches to reduce the parameter bit-width for lower storage and computation complexity. The main challenge of this type of approach is to maintain model accuracy. The methods can be further divided into **uniform quantization** and **non-uniform quantization** approaches.

*2.2.1 Uniform Quantization.* Some reduce bit-width to minimize computation and storage requirements. For instance, Gysel et al. [11] reduced the original 32-bit *weight* values to 8-bit floating-point values, following the IEEE-754 standard, and maintained reasonable model accuracy. However, even though the low bit-width floating-point numbers do require fewer resources, the floating-point computations still demand much more hardware resources and execution time as compared to fixed-point operations.

Therefore, some proposed to use the fixed-point format for further reduction of resource requirements. The conversion of floating-point *weight* values to fixed-point format is a quantization process. In general, the challenges for conversion to fixed-point is that the represented value range needs to be large enough to avoid the possibility of overflow while having a sufficiently fine resolution to minimize quantization errors.

One representative approach is by Lin et al. [15] that borrowed the idea from the mature signal processing field and optimized the signal-to-quantization-noise ratio (SQNR) of the mapping of the *weight* values to 8-bit fixed-point values. The authors reported only 0.6% model accuracy loss when applied this approach to AlexNet.

One challenge of the low bit-width fixed-point approach is that every CNN layer has a different value distribution, and the limited bit width cannot well cover the dynamic range of the whole network. To resolve this issue, Courbariaux et al. [20] used *dynamic-fixed-point* representation that allows a fixed-point format with a different *fractional part*, so that the customized representations can better cover each layer's data (*weight* and *activation*) values.

However, Guo et al. [24] reported severe model accuracy loss for bit-width less than 8. The accuracy loss of the low bit-width representation is mainly because a big chunk of near-zero data becomes indistinguishable zero values. Although some recent papers claim high accuracy results using 3-bit or 4-bit quantization of weights and activations [14] [33], they maintain the first and last layers in full precision. They hence are not suitable for the compact design purpose. Next, we discuss the *non-uniform* quantization approaches that provide a wider representation range.

*2.2.2 Non-uniform Quantization.* Han et al. [13] suggested that since the *weight* value distributions are not uniform, *non-uniform* quantization approaches with variable spacing between quantization points shall minimize model accuracy loss.

One widely adopted approach is the *logarithmic* quantization approach, which assigns the quantization points based on log-distribution. For this approach, the *weight* values are more evenly distributed across the quantization points, and the same number of bits can represent and differentiate a wider range than that by the *uniform* quantization approach.

Generally, the logarithmic quantization representation can more accurately represent most of the *weight* values. For instance, the heuristic Incremental Network Quantization (INQ) approach [17] took for each layer the maximum absolute value of each layer's *weight* value for the largest quantization point and set the remaining quantization points downward according to the log-scale, i.e., $2^i, 2^{i-1}, 2^{i-2}$..., with $2^i$ being the closest point to the maximum *weight* value.

In contrast, the LogQuant [25] and LogNet [29] assign to each layer with different quantization point distribution. One issue for hardware implementing of this method is that the different representation ranges on different layers will require additional overhead resources.

To address these issues, our proposed approach confines all data to the same [-1, 1] value range. In this way, we may use the same low bit-width representation for data and perform convolution computations of every layer. Details of our approach are elaborated in the following section.

## 3 Method

The main objective of our proposed approach is to produce a very compact CNN processor design. The key idea is having the same low bit-width representation throughout all inference layers without additional alignment process while maintaining model accuracy. We elaborate on our approach next.

### 3.1 Notations

We first define the notations used in this paper. Suppose that a full-precision (i.e., 32-bit floating-point) CNN model performs a series of matrix operations by dot products in each layer $l$, $1 \leq l \leq L$, with $L$ denoting the number of layers in the model. Then for each layer $l$ we have

$$Z^{[l]} = W^{[l]} * A^{[l-1]} + b^{[l]}, \qquad (1)$$

$$A^{[l]} = g^{[l]}(Z^{[l]}), \qquad (2)$$

where $W^{[l]}$ denotes the *weight* matrix, and $A^{[l-1]}$ is the input *activation* matrix of layer $l$. Note that $A^{[0]}$ is the first-layer input data, which can be an image or a segment of audio for analysis. Additionally, for layer $l$, $b^{[l]}$ is the *bias* vector, and $g^{[l]}(.)$ is a non-linear *activation* function (e.g., ReLU). The intermediate output result $Z^{[l]}$ is passed through the non-linear *activation* function $g^{[l]}(.)$ to produce the *activation* matrix $A^{[l]}$ for the next layer. Generally, for a CNN model, the *weight* matrix $W^{[l]}$ can be a 4-dimensional matrix for the convolution layer or a 2-dimensional matrix for the fully-connected layer. Since all convolutional and fully-connected layers are to adopt the same computing process, to ease later discussion, we assume that all *weight* matrices are 2-dimensional $N*N$ matrices and the matrix component of the $l$-th layer is indexed as $W^{[l]}(i, j)$ for $1 \leq i, j \leq N$.

### 3.2 Normalization-based Logarithmic Computing

Next, we present the insights of our approach, describe its key components, and comprehensively investigate the implementation. We first investigate how to normalize full-precision activation and weight values to the range of [-1, 1], how each layer should conduct the computations, and then devise a computing kernel suitable for hardware implementations.

*3.2.1 Weight Normalization.* Now, for the layer $l$, to normalize the range of *weight* values to [-1, 1], we first select a *weight* normalization factor $f_W^{[l]}$, to compute the normalized *weight* matrix $\overline{W}^{[l]}$ for logarithmic quantization as the following for each layer,

$$\overline{W}^{[l]} = W^{[l]}/f_W^{[l]} \quad \text{or} \quad W^{[l]} = \overline{W}^{[l]}f_W^{[l]}, \qquad (3)$$

where $\overline{W}^{[l]}$ represents the normalized *weight* matrix and $f_W^{[l]}$ the weight normalization factor.

*3.2.2 Activation Normalization.* Similarly, we also normalize activations of all layers to [-1, 1] range. Assuming that the first layer of input data $A^{[0]}$ has already been normalized during preprocessing, we use the *activation* normalization factor $f_A^{[l]}$ to normalize the *activation* $A^{[l]}$. In practice, we may take the absolute maximum of the $A^{[l]}$ to be the normalization factor $f_A^{[l]}$, hence

$$\overline{A}^{[l]} = A^{[l]}/f_A^{[l]} \quad \text{or} \quad A^{[l]} = \overline{A}^{[l]}f_A^{[l]} \qquad (4)$$

Since we normalize all the *weight* and *activation* values to the same range [-1, 1], and require no data-alignment during the operations, the kernel complexity is significantly reduced. Nevertheless, after the kernel computations, the output *activation* distribution of each layer may deviate from the target value range. Therefore, we need to rescale the *activation* values of each layer back to [-1, 1] range by a layer-wise scaling factor $f^{[l]}$. Next, we show how to derive the scaling number.

Based on Eq. (1), Eq. (2), Eq. (3) and Eq. (4) we have

$$\overline{A}^{[l]}f_A^{[l]} = A^{[l]} \qquad \text{from (4)}$$
$$= g^{[l]}(Z^{[l]}) \qquad \text{from (2)}$$
$$= g^{[l]}(W^{[l]} * A^{[l-1]} + b^{[l]}) \qquad \text{from (1)}$$
$$= g^{[l]}\left(\overline{W}^{[l]}f_W^{[l]} * \overline{A}^{[l-1]}f_A^{[l-1]} + b^{[l]}\right) \qquad \text{from (3), (4)}$$

$$= g^{[l]}\left(\left[\overline{W}^{[l]} * \overline{A}^{[l-1]} + \frac{b^{[l]}}{f_W^{[l]} f_A^{[l-1]}}\right] * (f_W^{[l]} f_A^{[l-1]})\right). \quad (5)$$

If we let $\overline{b}^{[l]} = b^{[l]} / f_W^{[l]} f_A^{[l-1]}$, and re-adjust the equation, then we have

$$\overline{A}^{[l]} = g^{[l]}\left(\left[\overline{W}^{[l]} * \overline{A}^{[l-1]} + \overline{b}^{[l]}\right] * (f_W^{[l]} f_A^{[l-1]})\right) / f_A^{[l]}, \quad (6)$$

if $g^{[l]}(.)$ is a homogeneous function of degree 1, i.e. $g^{[l]}(\alpha x) = \alpha g^{[l]}(x)$, for all $\alpha$. We can further simplify the equation by introducing a scaled per-layer layer output,

$$\overline{Z}^{[l]} = \overline{W}^{[l]} * \overline{A}^{[l-1]} + \overline{b}^{[l]} \quad (7)$$

Then by having the scaling factor $f^{[l]} = f_W^{[l]} f_A^{[l-1]} / f_A^{[l]}$ for layer $l$, we have the normalized output *activations*,

$$\overline{A}^{[l]} = g^{[l]}(\overline{Z}^{[l]}) * f^{[l]} \quad (8)$$

In practice, we restrict the scaling factor $f^{[l]}$ to be of the form $2^i$, and therefore only one bit-shift operation is needed to calculate the output of each layer.

Note that if the $g^{[l]}(.)$ is a non-homogeneous function, or homogeneous function of degree greater than one, then we shall do two-step scaling, with two scaling factors,

$$f_1^{[l]} = f_W^{[l]} f_A^{[l-1]}, \quad (9)$$

and

$$f_2^{[l]} = f_A^{[l]}. \quad (10)$$

Then,

$$\overline{A}^{[l]} = g^{[l]}(\overline{Z}^{[l]} * f_1^{[l]}) / f_2^{[l]}, \quad (11)$$

Then, the two-step scaling can be achieved by two bit-shift operations as we constrain the scaling factors to be of the form $2^i$. To simplify our later discussions, we assume that $g^{[l]}(.)$ is a homogeneous function of degree 1.

### 3.3 Value Quantization

After the above scaling process, we have all weight and activation values confined in [-1, 1] range. We then use the following quantization methods to convert the value to low bit-width representation.

We assume that $w$ is one of the components in $\overline{W}$. Then we extract the rounded exponent value $e(w)$ and the sign of $w$, i.e., $s(w)$, as the following,

$$e(w) = -Round(log_2(|w|)), \quad (12)$$

$$s(w) = sign(w). \quad (13)$$

The function *Round(.)* rounds a number to its nearest integer number.

Now we assume that $b$ bits are used for the logarithmic representation of weight values, and one of the $b$ bits is used to represent the sign of the *weight* value. Hence, we may use the $b$-$1$ bits to represent the absolute exponent value part and default the exponent value to be negative. Since the $b$-$1$ bits can represent values 0 to $2^{(b-1)} - 1$, if we let $B = 2^{(b-1)} - 1$, the normalized logarithmic quantized value shall be one of the values $\pm 2^0, \pm 2^{-1}, \ldots, \pm 2^{-B}$.

With $e(w)$, $s(w)$ and $B$, we have the logarithmic quantization *weight* value,

$$\widehat{w} = Q(w) = \begin{cases} 2^{-B} * s(w), & e(w) > B; \\ 2^0 * s(w), & e(w) \leq 0; \\ 2^{-e(w)} * s(w), & otherwise, \end{cases} \quad (14)$$

Practically, we find that b=4 produces very good model accuracy. For the *activations*, we find $k=8$ is good for all the cases we tested. We now assume that $a$ is one of the components in $\overline{A}$, we define the nearest-fixed-point-value function $N(.)$ like the following,

$$N(a) = Round(a * 2^{k-1}) * 2^{-(k-1)}, \quad (15)$$

where, $(k-1)$ is the number of fraction bits, and the *Round* function rounds a number to an integer. With $N(a)$ and $s(a)$, the $Q_{Fixed_8}(.)$ is defined as the following,

$$\hat{a} = Q_{Fixed_k}(a) = \begin{cases} 0, & |a| \leq 2^{-k}; \\ (1 - 2^{-k-1}) * s(a), & |a| \geq 1 - 2^{-k-1}; \\ N(a), & otherwise, \end{cases} \quad (16)$$

The scaled *bias* $\overline{b}^{[l]}$ can be very small values. Therefore, we use double bit-width to represent *bias*, i.e.

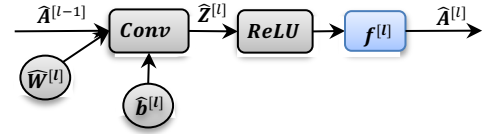

**Fig. 3**: A complete quantized execution of layer $l$. Assume that the *activation* function $g(.)$ is *ReLU*. The blue box performs scaling with the scaling factor $f^{[l]}$.

$$\hat{b}^{[l]} = Q_{Fixed_{16}}(\overline{b}^{[l]}) \quad (17)$$

Based on the above equations, we have a complete execution process of each layer shown in Fig. 3. Note that after scaling operation, all the output *activation* $\hat{A}^{[l]}$ is automatically scaled to the range [-1, 1], ready for the next layer's execution.

### 3.4 Training for Logarithmic Computing

The key difference of our training approach is to include the scaling factor into the training process and constrain all weight and activation values in the range of [-1, 1] following the architecture shown in Fig. 3. We adopt the CNN back-propagation training algorithm from [23] to calculate the gradient of the softmax loss function for all the *weights* in the network. We also adopt the approach proposed in [28] and quantize all parameters only during the forward propagation step. For each minibatch, the softmax cost C is calculated according to the network result and input targets, and the *weight* and *bias* are updated based on back-propagated gradients $gr_w$, $gr_b$ and $gr_f$. The algorithm is listed below.

| Algorithm 1: CNN training iteration for logarithmic computing |
|---|

**Input**: A minibatch of input data and targets $(\hat{A}^{[0]}, Y)$, model parameters and learning rate $\eta$.
**Output**: Updated parameters
{1. Computing the parameter gradients:}
{1.1 Forward propagation:}
1:    **for** $l = 1$ to $L$ **do**
2:       $\widehat{W}^{[l]} = Q(\overline{W}^{[l]})$
3:       $\hat{b}^{[l]} = Q_{Fixed_{16}}(\overline{b}^{[l]})$
4:       $\hat{f}^{[l]} = Q(\overline{f}^{[l]})$

```
5:      $\overline{A}^{[l]} = g^{[l]}(\widehat{A}^{[l-1]} * \widehat{W}^{[l]} + \widehat{b}^{[l]}) * \widehat{f}^{[l]}$
6:      $\widehat{A}^{[l]} = Q_{Fixed_8}(\overline{A}^{[l]})$
{1.2. Backward propagation:}
Compute
$[gr^w_{\widehat{A}^{[L]}}, gr^b_{\widehat{A}^{[L]}}, gr^f_{\widehat{A}^{[L]}}] = \frac{\partial C}{\partial \widehat{A}^{[L]}}$ based on $\widehat{A}^{[L]}$ and the target $Y$
7:      for $l = L$ to 1 do
8:          $gr^w_{\widehat{A}^{[l-1]}} = gr^w_{\widehat{A}^{[l]}} \widehat{W}^{[l]}$
9:          $gr_{w^{[l]}} = gr^{w\ T}_{\widehat{A}^{[l]}} * \widehat{A}^{[l-1]}$
10:         $gr^b_{\widehat{A}^{[l-1]}} = gr^b_{\widehat{A}^{[l]}} \widehat{b}^{[l]}$
11:         $gr_{b^{[l]}} = gr^{b\ T}_{\widehat{A}^{[l]}} * \widehat{A}^{[l-1]}$
12:         $gr^f_{\widehat{A}^{[l-1]}} = gr^f_{\widehat{A}^{[l]}} \widehat{b}^{[l]}$
13:         $gr_{f^{[l]}} = gr^{f\ T}_{\widehat{A}^{[l]}} * \widehat{A}^{[l-1]}$
{2. Update the un-quantized values from gradients:}
14:     for $l = 1$ to $L$ do
15:         $\overline{W}^{[l]} = (\widehat{W}^{[l]} - gr_{w^{[l]}} * \eta)$
16:         $\overline{b}^{[l]} = (\widehat{b}^{[l]} - gr_{b^{[l]}} * \eta)$
17:         $\overline{f}^{[l]} = (\widehat{f}^{[l]} - gr_{f^{[l]}} * \eta)$
```

## 4 Experiments

In this section, we evaluate the impact of low bit-width on model accuracy. We conducted a large number of tests on the large-scale classification of ImageNet, which is the most challenging image classification benchmark. The ImageNet dataset has approximately 1.2 million training images and 50,000 validation images. Each image in the dataset is assigned a label in one of the 1000 classes. In all experiments, we take the full-precision CNN models from the *PyTorch* and perform quantization to obtain the initial solutions for our training. We conduct training on the models under the batch size of 512 (for AlexNet), 256 (for VGG16), and 512/256 (for ResNet-18/34). During the training, we set the initial learning rate as 0.0002, which continues to decrease to 10 times smaller every ten epochs. We adopt a stochastic gradient descent optimizer with *weight* decay equals 0.0001, and the momentum equals 0.9. Experiments are conducted on Linux machines with Intel CPUs and NVIDIA TESLA P100 graphic processing units.

For the sake of brevity, we introduce the symbol x/y to represent the bit-width of the *weight* **W** and the bit-width of the *activation* **A** for quantization. Besides, we also use 'f' to indicate a 32-bit full precision floating-point. For instance, 4/f represents the 4-bit *weight* and 32-bit full precision *activation*.

### 4.1 Image Classification Results

We first verify our approach on the 10-class number recognition LeNet-5 network [27] with 4-bit *weight* and 8-bit *activation*, and fully maintain the same model accuracy.

Next, we test on AlexNet, the winner of ILSVRC 2012, which has five convolutional layers and three fully-connected layers. The second network tested is VGG16, which has 16 layers, including three fully-connected layers, and 138 million parameters. Compared with AlexNet, VGG16 requires much more arithmetic operations while achieving a higher prediction accuracy. Third, we test ResNet. Different from the above networks, ResNet resolves the vanishing gradient problem using shortcut connections. We first test the 18-layer version for exploration purposes and then the 34-layer version. Finally, we apply the ImageNet validation dataset to evaluate the accuracy of our quantized networks.

Table 1 summarizes the model accuracy results of our low bit-width 4/8 quantized networks and the full-precision reference networks. Generally, our low bit-width networks have only negligible (less than 1%) model accuracy loss.

Table 1: Model accuracy comparison with full-precision networks.

| Network | Param. | Top-1(%) | Top-5(%) |
|---|---|---|---|
| **AlexNet ref** | f/f | 56.52 | 79.56 |
| ours | 4/8 | 56.324 | 79.226 |
| **VGG16 ref** | f/f | 71.592 | 90.392 |
| ours | 4/8 | 71.132 | 90.37 |
| **ResNet18 ref** | f/f | 69.758 | 89.778 |
| ours | 4/8 | 69.176 | 89.538 |
| **ResNet34 ref** | f/f | 73.196 | 91.73 |
| ours | 4/8 | 73.02 | 91.618 |

### 4.2 Compare with Other Quantization Methods

Our approach also produces results with minimum accuracy loss when compared with other quantization approaches on AlexNet, VGG16, and ResNet-18, as summarized in Table 2. Our networks not only use the least number of bits but also have less than 0.5% negligible model accuracy loss. Outstandingly, our 4/8 networks require 87.5% less of *weight* data size and 75% less of the *activation* data size. Although INQ [17] produces a slightly higher accuracy for AlexNet, it uses full precision *activation* and requires a much higher computation and storage cost. Although the ResNet-18 has a more compact model than other models, with very few redundant *weights*, our method can still maintain top model accuracy under low bit-width constraints.

Table 2: Comparison with different approaches.

| Network | Param | Top-1 | Top-5 |
|---|---|---|---|
| **AlexNet ref** | f/f | 56.52 | 79.56 |
| INQ[17] | 5/f | 57.39 | 80.46 |
| Ristretto[11] | 8/8 | 53.57 | 78.25 |
| LogQuant[25] | 5/4 | - | 70.6 |
| Ours | 4/8 | 56.32 | 79.23 |
| **VGG16 ref** | f/f | 71.592 | 90.39 |
| INQ [17] | 5/f | 70.82 | 90.3 |
| LogNet [29] | 4/f | - | 85.2 |
| Ours | 4/8 | 71.13 | 90.37 |
| **ResNet-18 ref** | f/f | 69.76 | 89.78 |
| INQ[17] | 5/f | 68.89 | 89.10 |
| Ours | 4/8 | 69.18 | 89.54 |

### 4.3 Object detection results

To extensively study the capabilities of the proposed techniques, we evaluated more complex vision tasks, comparing the mean Average Precision (mAP) as a measure of the 4/8 bit-width implementation and the full-precision of Yolov2 model on the MS-COCO dataset [30] for object detection.

Yolov2 is one popular object detection method. It performs one of the best tradeoffs between the accuracy and the inference speed for object detection. Compared to other object detection networks, it uses a single neural network to predict object bounding boxes and class probabilities in a single assessment, so it is more suitable for execution in embedded systems. MS-COCO is the most popular object detection dataset with 80 categories. It is widely used to benchmark object detectors because of its rich annotations and challenging scenarios.

Table 3: Model accuracy comparison of Yolov2

| Network | Param. | mAP (%) |
|---|---|---|
| **Yolov2 ref** | f/f | 49.5 |
| ours | 4/8 | 47.9 |

As we all know, an object detection task is challenging to process using fewer bits because its complexity is much larger than the classification task. Table 3 shows our low bit-width 4/8 Yolov2 network has merely 1.6 % lower mAP than the full-precision reference version.

### 4.4 FGPA Implementation

We also perform stress tests by implementing our 4/8 Yolov2 network onto Xilinx Zynq 7z020, which has about 75% less capacity than 7z045. The comparison results with other FPGA-based implementations are summarized in Table 4. Zhao [26] implemented a 32-bit fixed-point Yolov1, which has six fewer layers and 1/4 activations than Yolov2, on the 7z045, which has about four times the capacity of 7z020. Our work achieves 21% less latency and near 2.56 times better throughput against Zhao's. Our 4/8 Tiny-Yolov2 on Xilinx 7z020 achieves 1.38 times better throughput than Wai's 16-bit fixed-point Intel Cyclone V implementation [21].

Table 4: Comparison of FPGA-based CNN processors.

|  | Zhao [26] | Wai [21] | our work | |
|---|---|---|---|---|
| Model | Yolov1 | Tiny-Yolov2 | Yolov2 | Tiny-Yolov2 |
| FPGA | ZC706 Zynq 7z045 | Cyclone V PCIe | PYNQ-Z1 Zynq 7z020 | |
| Clock Rate | 200 MHz | 117 MHz | 200 MHz | |
| Bit-width | 32 | 16 | 4/8 | 4/8 |
| Weights (MB) | 228.85 | 21.41 | 25.35 | 5.35 |
| GOPs/Byte | 130.79 | 146.62 | 638.54 | 421.6 |
| Logic | - / 218K | - / (64%) | 49,158 / 53,200 (92%) | |
| Block RAM | - / 545 | / (40%) | 104 / 140(74%) | |
| DSP | - / 900 | / (41%) | 124 / 280 (56%) | |
| Latency (s) | 0.744 | 0.278 | 0.611 | 0.202 |
| GOPS | 18.82 | 19.42 | 48.23 | 26.73 |

### 4.5 IC Implementation

After thoroughly verifying our processor design on FPGA, we also conduct an IC design for Yolov2 using TSMC 40 nm cell library. Our 4/8 networks have 128 processing elements and take only 0.15 mm$^2$ silicon area, a truly compact size. The design is underway to be integrated into an SoC for embedded applications.

## 5 Conclusion

In this paper, we modify the normalization-based logarithmic computing scheme that significantly reduces the convolution kernel design. We apply our method to several popular CNN architectures such as AlexNet, VGG16, and ResNet, and conduct extensive experiments on the ImageNet dataset. Besides, we also apply our method to the Yolov2 model for the object detection task. The hardware implementation is very compact and efficient and is perfect for edge computing.

## REFERENCES

[1] Luo, Jian-Hao, Jianxin Wu, and Weiyao Lin. "Thinet: A filter level pruning method for deep neural network compression." Proceedings of the IEEE international conference on computer vision. 2017.
[2] Sahin, Suhap, Yasar Becerikli, and Suleyman Yazici. "Neural network implementation in hardware using FPGAs." International Conference on Neural Information Processing. Springer, Berlin, Heidelberg, 2006.
[3] Gong, Yunchao, et al. "Compressing deep convolutional networks using vector quantization." arXiv preprint arXiv:1412.6115 (2014).
[4] Krizhevsky, Alex, Ilya Sutskever, and Geoffrey E. Hinton. "Imagenet classification with deep convolutional neural networks." Advances in neural information processing systems. 2012.
[5] Simonyan, Karen, and Andrew Zisserman. "Very deep convolutional networks for large-scale image recognition." arXiv preprint arXiv:1409.1556 (2014).
[6] Zhu, Chenzhuo, et al. "Trained ternary quantization." arXiv preprint arXiv:1612.01064 (2016).
[7] Rastegari, Mohammad, et al. "Xnor-net: Imagenet classification using binary convolutional neural networks." European conference on computer vision. Springer, Cham, 2016.
[8] Gysel, Philipp, Mohammad Motamedi, and Soheil Ghiasi. "Hardware-oriented approximation of convolutional neural networks." arXiv preprint arXiv:1604.03168 (2016).
[9] Zhou, Shuchang, et al. "Dorefa-net: Training low bitwidth convolutional neural networks with low bitwidth gradients." arXiv preprint arXiv:1606.06160 (2016).
[10] Courbariaux, Matthieu, et al. "Binarized neural networks: Training deep neural networks with weights and activations constrained to+ 1 or-1." arXiv preprint arXiv:1602.02830 (2016).
[11] Gysel, Philipp, et al. "Ristretto: A framework for empirical study of resource-efficient inference in convolutional neural networks." IEEE transactions on neural networks and learning systems 29.11 (2018): 5784-5789.
[12] He, Yihui, Xiangyu Zhang, and Jian Sun. "Channel pruning for accelerating very deep neural networks." Proceedings of the IEEE International Conference on Computer Vision. 2017.
[13] Han, Song, Huizi Mao, and William J. Dally. "Deep compression: Compressing deep neural networks with pruning, trained quantization and huffman coding." arXiv preprint arXiv:1510.00149 (2015).
[14] Choi, Jungwook, et al. "Pact: Parameterized clipping activation for quantized neural networks." arXiv preprint arXiv:1805.06085 (2018).
[15] Lin, Darryl, Sachin Talathi, and Sreekanth Annapureddy. "Fixed point quantization of deep convolutional networks." International conference on machine learning. 2016.
[16] Denil, Misha, et al. "Predicting parameters in deep learning." Advances in neural information processing systems. 2013.
[17] Zhou, Aojun, et al. "Incremental network quantization: Towards lossless cnns with low-precision weights." arXiv preprint arXiv:1702.03044 (2017).
[18] Dettmers, Tim. "8-bit approximations for parallelism in deep learning." arXiv preprint arXiv:1511.04561 (2015).
[19] LeCun, Yann, John S. Denker, and Sara A. Solla. "Optimal brain damage." Advances in neural information processing systems. 1990.
[20] Courbariaux, Matthieu, Yoshua Bengio, and Jean-Pierre David. "Training deep neural networks with low precision multiplications." arXiv preprint arXiv:1412.7024 (2014).
[21] Wai, Yap June, et al. "Fixed point implementation of tiny-yolo-v2 using opencl on fpga." International Journal of Advanced Computer Science and Applications 9.10 (2018): 506-512.
[22] Ioffe, Sergey, and Christian Szegedy. "Batch normalization: Accelerating deep network training by reducing internal covariate shift." arXiv preprint arXiv:1502.03167 (2015).
[23] Rumelhart, David E., Geoffrey E. Hinton, and Ronald J. Williams. "Learning representations by back-propagating errors." nature 323.6088 (1986): 533-536.
[24] Guo, Kaiyuan, et al. "[DL] A survey of FPGA-based neural network inference accelerators." ACM Transactions on Reconfigurable Technology and Systems (TRETS) 12.1 (2019): 1-26.
[25] Miyashita, Daisuke, Edward H. Lee, and Boris Murmann. "Convolutional neural networks using logarithmic data representation." arXiv preprint arXiv:1603.01025 (2016).
[26] Zhao, Ruizhe, et al. "Optimizing CNN-based object detection algorithms on embedded FPGA platforms." International Symposium on Applied Reconfigurable Computing. Springer, Cham, 2017.
[27] LeCun, Yann, et al. "Gradient-based learning applied to document recognition." Proceedings of the IEEE 86.11 (1998): 2278-2324.
[28] Courbariaux, Matthieu, et al. "Binarized neural networks: Training deep neural networks with weights and activations constrained to+ 1 or-1." arXiv preprint arXiv:1602.02830 (2016).
[29] Lee, Edward H., et al. "Lognet: Energy-efficient neural networks using logarithmic computation." 2017 IEEE International Conference on Acoustics, Speech and Signal Processing (ICASSP). IEEE, 2017.
[30] Lin, Tsung-Yi, et al. "Microsoft coco: Common objects in context." European conference on computer vision. Springer, Cham, 2014.
[31] Zhao, Ruizhe, et al. "Optimizing CNN-based object detection algorithms on embedded FPGA platforms." International Symposium on Applied Reconfigurable Computing. Springer, Cham, 2017.
[32] Salimans, Tim, and Durk P. Kingma. "Weight normalization: A simple reparameterization to accelerate training of deep neural networks." Advances in neural information processing systems. 2016.
[33] Baskin, Chaim, et al. "Nice: Noise injection and clamping estimation for neural network quantization." arXiv preprint arXiv:1810.00162 (2018).